\IfFileExists{filecontents.sty}{\RequirePackage{filecontents}}{}

\documentclass[11pt,a4paper]{article}
\usepackage{authblk}
\usepackage[hyperref]{acl2017}
\usepackage{times}
\usepackage{latexsym}
\usepackage[inline]{enumitem}
\usepackage{linguex}
\usepackage{url}
\usepackage{comment}
\usepackage{blindtext}

\usepackage{booktabs}

\usepackage[]{amsmath}
\usepackage[]{amsthm}
\usepackage{amstext}
\usepackage{amssymb}
\usepackage{todonotes}
\usepackage{hhline}
\usepackage{multirow}

\newcommand{\fakeparagraph}[1]{\vspace{1mm}\noindent\textbf{#1.}~~}

\aclfinalcopy 

\title{
Word Sense Disambiguation with LSTM:\\Do We Really Need 100 Billion Words?
}

\author{Minh Le$^1$, Marten Postma$^1$, Jacopo Urbani$^2$ \\
$^1$ Department of Language, Literature and Communication, Vrije Universiteit Amsterdam \\
$^2$ Department of Computer Science, Vrije Universiteit Amsterdam \\
{\tt \{m.n.le,m.c.postma\}@vu.nl}, {\tt jacopo@cs.vu.nl}
}
\date{}

\begin{document} \maketitle \begin{abstract} Recently,
    \citeauthor{Yuan2016}~\shortcite{Yuan2016} have shown the effectiveness of
    using Long Short-Term Memory (LSTM) for performing Word Sense Disambiguation
    (WSD). Their proposed technique outperformed the previous state-of-the-art
    with several benchmarks, but neither the training data nor the source code
    was released. This paper presents the results of a reproduction study of
    this technique using only openly available datasets (GigaWord, SemCore, OMSTI)
    and software (TensorFlow). From them, it emerged that state-of-the-art
    results can be obtained with much less data than hinted by
    \citeauthor{Yuan2016} All code and trained models are made freely available.
\end{abstract}

\section{Introduction}

Word Sense Disambiguation (WSD) is a long-established task in the NLP
community~\cite{navigli2009word}.  Many approaches have been proposed -- the
more popular ones include the usage of Support Vector Machine (SVM)
\cite{Zhong:2010:MSW:1858933.1858947}, combined with unsupervisedly-trained
embeddings \cite{P16-1085,P15-1173}, and graph-based approaches
\cite{agirre2014random}, which can also be combined with embeddings
\cite{tripodi2016game}.

Recently, \newcite{Yuan2016} coupled a Long short-term memory (LSTM) language
model~\cite{hochreiter.schmidhuber97} trained on 100B words with small
sense-annotated corpora to achieve state-of-the-art performance in all words
WSD. This method was designed to address, in a distant supervision fashion, what
is believed to be one central problem in WSD, that is ``lack of sufficient
labeled training data''~(\newcite{Yuan2016}, first page). This method addresses
this problem by training a language model with an LSTM network and  a huge
\emph{unlabeled} textual corpus of 100B words, and then construct sense
embeddings using a much smaller \emph{labeled} dataset.

Even though the results obtained by~\citeauthor{Yuan2016} outperform the
previous state-of-the-art, this work has two important limitations:
\emph{First}, neither the datasets nor the models are available to the community
and, to the best of our knowledge, there is no other publicly available dataset
which comes close to the size of the one used in the work. This is unfortunate
because this unavailability makes the re-application of this technique in other
contexts a non-trivial process, especially because the input size was indicated
as one of the main reason behind the excellent performance. \emph{Second}, it is
unclear which are the problems that prevent even higher accuracies. These
problems could be algorithmic, relate to the input (either size or quality), or
both. Identifying the nature of these problems is crucial in order to limit the
search space for further improvements.

To address these two issues, we reimplemented \citeauthor{Yuan2016}'s method with
the goal of reproducing and make available the results. While a full replication
is not possible due to the unavailability of the original data, our work allowed
us to perform a deeper investigation on the actual performance of this
technique. This allows us to provide a first, preliminary, answer to the
following two questions:

\begin{itemize}

\item \textit{How sensitive to the amount of training data is a WSD process
    driven by a large-scale statistical model? In other words, do we really need very large
training sets to achieve state-of-the-art performance?}

\item \textit{Are there some limitations that cannot be overcome by using larger
    training sets?}

\end{itemize}

The contribution of this paper is thus two-fold: On the one side, we present a
reproducibility study whose results are publicly available and hence can be
freely used by the community. Notice that the lack of available models has been
explicitly mentioned, in a recent work, as the cause for the missing comparison
of this technique with other competitors~\cite[footnote 10]{raganato}.  On the
other side, we investigate on the effective role played by the training set, and
report our conclusions to shed more light into the value of this and similar
methods.

We anticipate one interesting conclusion that emerged from our analysis, which
is that we were
able to replicate the same performance as reported in~\newcite{Yuan2016}, but
using only a corpus of 1.8B words (GigaWords), which is less than 2\% of the
data used in the original study. 

In the remaining, we proceed as follows:
First, we describe the original work in Section~\ref{sec:model}. Then,
we report the methodology for our reproduction study and other technical details
in Section~\ref{sec:methodology}. The results and related analysis are reported in
Section~\ref{sec:results}. 
Section~\ref{sec:conclusions} concludes the paper.

\section{WSD with Language Models}
\label{sec:model}


Broadly speaking, the method proposed by~\citeauthor{Yuan2016} performs WSD by:
1) constructing a language model from a large textual corpus; 2) extracting
sense embeddings from this model using a much smaller annotated corpus; 3) rely
on the sense embeddings to make predictions in unseen sentences. Each operation
is described below.

\subsection{Constructing the language model}
\label{sub:lstm}

The first operation consists of constructing a language model that produces
embeddings for phrases using a Long Short-Term Memory
(LSTM)~\cite{hochreiter.schmidhuber97}. LSTM is a celebrated recurrent neural
network architecture that has proven effective in many natural language
processing tasks \cite[among others]{sutskever2014sequence,Dyer2015,He2017}.
Different from previous networks, LSTM is equipped with trainable gates that
control the flow of information, allowing the neural networks to learn both
short- and long-range dependencies (we refer the interested reader to the
textbook by~\citeauthor{Goodfellow-et-al-2016}
and~\cite{hochreiter.schmidhuber97} for more details on this network).





\begin{figure}[t]
\centering
\includegraphics[width=0.4\textwidth]{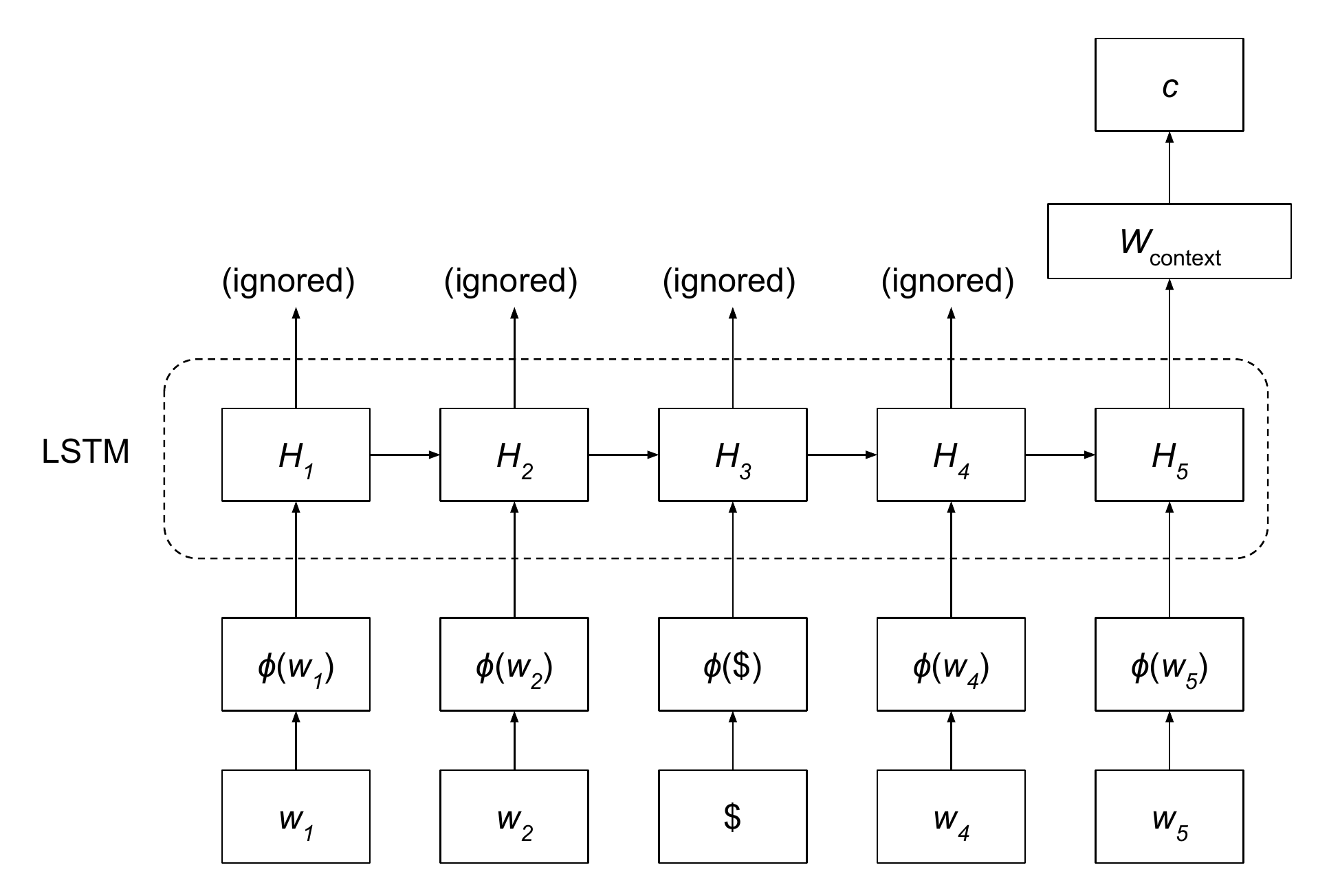}
\caption{The LSTM model used to perform language modeling and compute context embeddings.}
\label{fig:lstm}
\end{figure}

In \newcite{Yuan2016}, an LSTM neural network is used to capture the meaning of
words in context.  Given a sentence $s=(w_1, w_2,\dots, w_n)$, they replace word
$w_k\,(1 \le k \le n)$ by a special token $\$$.  The model takes this new
sentence as input and produces a context vector ${\mathbf{c}}$ of dimensionality
$p$ (see Figure~\ref{fig:lstm}).\footnote{As usual, vectors are indicated
with boldface to distinguish them to scalar and other symbols.}



Each word $w$ in the vocabulary $\mathcal{V}$ is associated with an (output) embedding $\phi_{\mathrm{o}}(w)$ of the same
dimensionality.\footnote{Notice that each word has one more vector used as input
to the neural network. Those vectors are separate from the output embeddings.}
The model is trained to predict the omitted word,
minimizing the softmax loss $\ell$ over a big collection of sentences $\mathcal{D}$:

\vspace{-0.5em}
\[\ell = - \sum_{s \in \mathcal{D}} \sum_{k=1}^{|s|} \log
\frac{\exp(\mathbf{c} \cdot {\phi_{\mathrm{o}}(w_k)})}
{\sum_{w^\prime \in \mathcal{V}} \exp(\mathbf{c} \cdot {\phi_{\mathrm{o}}(w^\prime)})} \]

After the model is trained, we can use it to extract \emph{context} embeddings,
i.e., latent numerical representations of the sentence
surrounding a given word.

\subsection{Calculating the sense embeddings}
\label{sec:sense-emb}

The model produced by the LSTM network is meant to capture the ``meaning'' of
words in the context they are mentioned. In order to perform the sense
disambiguation, we need to extract from it a suitable representation for
word senses. To this purpose, the method relies on another corpus where each
word is annotated with the corresponding sense.


The main intuition is that words intended with the same sense are mentioned in
contexts which are very similar to each other. This suggests a simple way to
calculate sense embeddings. First, the LSTM model is invoked to compute the
context vector for each occurrence of one sense in the annotated dataset. Once
all context vectors are computed, the sense embedding is defined as the average
of all vectors. Let us assume, for instance, that the sense $horse_n^2$ (that
    is, the second sense of horse as a noun) appears in the two
    sentences:

\ex. The move of the $horse_n^2$ to the corner forced the checkmate.

\ex. Karjakin makes up for his lost bishop a few moves later, trading rooks and
winning black's $horse_n^2$.

In this case, the method will replace the sense by $\$$ in the sentences and
feed them to the trained LSTM model to calculate two context vectors
${\mathbf{c_1}}$ and ${\mathbf{c_2}}$. The sense
embedding $\mathbf{s_{horse_n^2}}$ is then computed as:

\[\mathbf{s_{horse_n^2}} = \frac{\mathbf{c_1} + \mathbf{c_2}}{2}\]

This procedure is computed for any sense that appears in the annotated corpus.


\subsection{Predicting the senses}

After all sense embeddings are computed, the method is ready to disambiguate
target words. This procedure proceeds as follows:

\begin{enumerate}

    \item Given an input sentence and a target word, it replaces
        the occurrence of the target word by $\$$ and uses the LSTM model to
        predict a context vector $\mathbf{c_t}$.

    \item The lemma of the target word is used to retrieve from Wordnet the candidate synsets
        $s_1,\ldots,s_n$ where $n$ is the number of synsets.
        Then, the procedure looks up the corresponding sense embeddings
        $\mathbf{s_1},\ldots,\mathbf{s_n}$ computed in
        Section~\ref{sec:sense-emb}.

    \item The procedure invokes a subroutine to choose one of the $n$ senses for the context vector
        $\mathbf{c_t}$. The paper proposes two implementations for this
        subroutine:
        \begin{enumerate}
            \item The procedure selects the sense whose vector is closest to
                $\mathbf{c_t}$ using the cosine similarity as distance function;
            \item The procedure invokes a label propagation algorithm to perform
                a more sophisticated classification where also non annotated
                sentences which contain the target word (picked from the large
            textual corpus), are taken into account.
        \end{enumerate}

\end{enumerate}

In our study, we attempted at implementing the procedure 3(b) but we encountered
technical difficulties we did not solved yet. Thus, we will limit to evaluate
procedure 3(a) for the sense prediction, and leave 3(b) as future work.

\section{Reproduction Study: Methodology}
\label{sec:methodology}

Before we report the results of our experiments, we describe the datasets we
used and give some details regarding our implementation.

\fakeparagraph{Training data} For the training of the LSTM model, we make use of
the English Gigaword Fifth Edition\footnote{Linguistic Data Consortium (LDC)
catalog number LDC2011T07}. The corpus consists of 1.8 billion tokens in 4.1
million documents, originated from four major news agencies. We leave the study
of bigger corpora for future work.

For the training of the sense embeddings, we use the same corpora used by Yuan
et al.:

\begin{enumerate}

    \item \emph{SemCor}~\cite{miller1993semantic} is a corpus containing approximately
        240,000 sense annotated words. The tagged documents originate from the
        Brown corpus \cite{francis1979brown} and cover various genres.

    \item \emph{OMSTI}~\cite{K15-1037} contains one million sense annotations
        automatically tagged by exploiting an the English-Chinese part of the
        parallel MultiUN corpus.

\end{enumerate}

\fakeparagraph{Implementation} We used the BeautifulSoup HTML parser to extract
the text from Gigawords. Then, we used the English
models\footnote{\texttt{en\_core\_web\_md-1.2.1}} of Spacy 1.8.2 for sentence
boundary detection and tokenization.



The LSTM model is implemented using TensorFlow 1.2.1 \cite{Abadi2015}. We chose TensorFlow
because of its industrial-grade quality and because it comes with vital
functionalities for training large-scale models.

The main bottleneck of Yuan et
al.'s approach is the training of the LSTM model. Although we do not use a
100-billion word corpus, training the model on Gigaword can already take
years if not optimized properly.

To reduce training time, we assumed that all (padded) sentences in the
batch have the same length. This optimization increases the speed by 17\%.
Second, following Yuan et al., we use sampled softmax loss \cite{Jean2014}.
Third, we grouped sentences of similar length together while varying the number
of sentences in a batch to fully utilize GPU memory.  Together, these heuristics
increased training speed by 42 times.

\fakeparagraph{Evaluation framework} For evaluating the WSD predictions, we
selected two test sets: one from Senseval
2~\cite{palmer-EtAl:2001:SENSEVAL}, which tests the disambiguation of nouns,
verbs, adjectives and adverbs, and one from SemEval 2013~\cite{navigli-jurgens-vannella:2013:SemEval-2013}, which focuses only on
nouns.

The test test from Senseval 2 is the English All-Words Task; \emph{senseval2}
henceforth. This dataset contains 2387 annotations from three articles from the
Wall Street Journal. Most of the annotations are nominal, but the competition
also contains annotations for verbs, adjectives, and adverbs. About 67\% of all
tokens are annotated with the most frequent sense (MFS). This means that the
simple and well-known heuristic would score 67\% of accuracy on this dataset.

The test set from SemEval 2013 is the one taken from the task 12: Multilingual
Word Sense Disambiguation; \emph{sem2013-aw} henceforth. This task consists
of two disambiguation tasks: Entity Linking and Word Sense Disambiguation for
English, German, French, Italian, and Spanish. We restricted ourselves to
the WSD part using WordNet as a sense repository. The test set contains 13
articles obtained from previous editions of the workshop on Statistical Machine
Translation.\footnote{\url{http://www.statmt.org}} The articles cover different
domains, ranging from sports to financial news. With respect to the English WSD
part, there are 1,644 test instances in total, all nouns. The sense repository
used to tag these instances is WordNet version 3.0. Based on this sense
repository, we computed the number of instances annotated with the MFS, and the
number of instances annotated with one of the least frequent sense (LFS). Out
of the 1,644 test instances, the MFS applies in 1,035 cases, while one of the
LFS applies in the rest of 609 cases. This gives a baseline of 62.96\% for the
MFS heuristic.

\section{Results}
\label{sec:results}

\begin{table}[t]
\centering \small
\begin{tabular}{lrrr}
\toprule
Vocab. &    $p$ &     $h$ & \#params \\
\midrule
    1M &   10 &   100 &     20M \\
    1M &   64 &   256 &    128M \\
    1M &  128 &   512 &    258M \\
    1M &  512 &  2048 &   1050M \\
\bottomrule
\end{tabular}
\caption{LSTM models used in our paper. $h$ is the number of LSTM units and $p$ is embedding dimensionality.}
\label{tab:models}
\end{table}

For training the LSTM-based language model, we have been using a high-end
machine equipped with a Intel Xeon E5, 256GB of RAM, and a nVIDIA GeForce GTX
1080 Ti with 12GB of RAM. We trained the network with different parameters (Table~\ref{tab:models}).
Despite the optimizations described before and the relatively recent
hardware, our implementation still requires about a day to finish an epoch with $h=2048$ hidden units and $p=512$ embedding dimensions (henceforth called the ``Google'' setting). At the time this paper is written,
the training is still ongoing and has reached the 180\textsuperscript{th} epoch, which means
the program is running for about 4.5 months.

In the following, we will first look at the WSD predictions produced by our
model and compare with the original work and other state-of-the-art models (Section~\ref{sec:wsd}).
Then, we report other experiments where we varied the amount of training data (Section~\ref{sec:effect-of-data})
and other parameters of the language model (Section~\ref{sec:capacity}).
Finally, we perform stability check to make sure that the conclusions hold when models are retrained (Section~\ref{sec:stability}).

\subsection{WSD predictions}
\label{sec:wsd}

\begin{table*}[th!]
\centering
\scriptsize
\begin{tabular}{lllllllllllll}
\toprule
      &      &  \multicolumn{5}{c}{senseval2} & \multicolumn{5}{c}{sem2013-aw} \\
Model & MFS back-off &      P &      R &     F1 &  R\_mfs &  R\_lfs &      P &      R &     F1 &  R\_mfs &  R\_lfs \\
\midrule
                Our implementation (T: SemCor) &           No &  0.700 &  0.671 &  0.685 &  0.808 &  0.392 &  0.675 &  0.641 &  0.658 &  0.825 &  0.327 \\
                Our implementation (T: SemCor) &          Yes &  0.700 &  0.700 &  0.700 &  0.850 &  0.392 &  0.666 &  0.666 &  0.666 &  0.866 &  0.327 \\
                 Our implementation (T: OMSTI) &           No &  0.694 &  0.455 &  0.549 &  0.541 &  0.278 &  0.696 &  0.489 &  0.574 &  0.614 &  0.278 \\
                 Our implementation (T: OMSTI) &          Yes &  0.674 &  0.674 &  0.674 &  0.867 &  0.278 &  0.655 &  0.655 &  0.655 &  0.876 &  0.278 \\
          Our impl.\ (T: SemCor+OMSTI) &           No &  0.682 &  0.653 &  0.667 &  0.781 &  0.392 &  0.677 &  0.642 &  0.659 &  0.798 &  0.378 \\
          Our impl.\ (T: SemCor+OMSTI) &          Yes &  0.682 &  0.682 &  0.682 &  0.823 &  0.392 &  0.668 &  0.668 &  0.668 &  0.839 &  0.378 \\
\midrule
 Yuan et al. (2016) (T: Semcor)* &            - &      - &      - &  0.736 &      - &      - &      - &      - &  0.670 &      - &      - \\
  Yuan et al. (2016) (T: OMSTI)* &            - &      - &      - &  0.724 &      - &      - &      - &      - &  0.673 &      - &      - \\
Raganato et al. (2017)* &            - &      - &      - &  0.720 &      - &      - &      - &      - &  0.669 &      - &      - \\
Weissenborn et al. (2015)* &            - &      - &      - &  - &      - &      - &      - &      - &  0.728 &      - &      - \\
Iacobacci et al. (2016)* &            - &      - &      - &  0.683 &      - &      - &      - &      - &  - &      - &      - \\
\bottomrule
\end{tabular}
\caption{Performance of our implementation compared to already published
results. A star (*) at the end of a model name indicates that the results are copied from the respective paper.}
\label{tab:results}
\end{table*}

Since our goal is to reproduce the results presented in the original work, we
intended to use with the ``Google'' setting which was reported as leading to the best performance. However, since the
training is still ongoing, we started to perform WSD using a model trained with
less iterations. To our surprise, already after the 65\textsuperscript{th} iteration we did
not observe any improvement in the downstream WSD task. This does not mean that
the training did not produce better models. Indeed, after this iteration the
training algorithm returned new models with a lower negative log likelihood in language modeling. However, none
of these models yielded better performance for WSD, thus we will be using the
model produced at the 65\textsuperscript{th} iteration to compare our results to the
original work.

Table~\ref{tab:results} presents the results. Notice that we report the
accuracy when either SemCor or OMSTI was used for creating the sense embeddings.
For each experiment, we report the results with and without a \emph{MFS back-off}
strategy. This strategy is concerned with the choice of the correct sense
whenever the annotated corpus has no evidence for it. In this case, if the
strategy is activated then we simply choose the most frequent sense according to WordNet \cite{Fellbaum-WordNet-1998}.
The results reveal that the heuristics is crucial to obtain high performance, especially when sense embeddings are trained on OMSTI.\footnote{\citeauthor{Yuan2016} did not report if they use a MFS back-off strategy or not.}

For each experiment we report \emph{Precision (P)},
\emph{Recall (R)}, and \emph{F1}.\footnote{See \citeauthor{navigli2009word}
(\citeyear{navigli2009word}) for the definitions of the metrics.} The recall is
also reported on two subsets of the test set: most-frequent-sense instances (\emph{R\_mfs}) and
less-frequent-sense instances (\emph{R\_lfs}). The first subset (MFS) contains all
test instances for which the answer is the MFS of a lemma, whereas the second
subset (LFS) contains all the remaining test instances.
Since the MFS is such a strong baseline in WSD, there is a tendency for WSD systems to overfit towards the MFS \cite{C16-1330}. A good performance on the recall on the LFS instances, while still having a high overall F1 score, is a good indicator of how balanced a system is with respect to its sense assignment.

We now compare our reimplemented model to \newcite{Yuan2016} and other state-of-the-art WSD systems.
When we use SemCor to train the synset embeddings, we come close to the results reported in the original publication (0.7 vs.\ 0.736 using senseval2 and 0.666 vs.\ 0.67 using sem2013-aw).
On senseval2, our model outperforms the model from \citet{P16-1085} and comes close to the results from \citet{raganato}.
On sem2013-aw, the gap with \citet{P15-1058} is still significant. When we use OMSTI for the annotated data,
the difference is a bit more pronounced (0.724 vs. 0.674 on senseval2 and 0.673
vs. 0.655 using sem2013-aw) but our model still beats many other systems in this competition.
Using both SemCor and OMSTI to train the synset embeddings does not seem to affect the results. The results in the \emph{R\_lfs} columns show that our model does not overfit towards the MFS. Whereas it is not uncommon for WSD systems to obtain 0.90 recall on MFS instances (\emph{R\_mfs}) and 0.25 on LFS instances \emph{R\_mfs} \cite{POSTMA16.437}, our sense assignment is much more balanced. Finally, all these results were obtained by training on a much smaller corpus. Gigawords is only 2\% than the (non-disclosed) dataset used in the original work and the training required only 65 iterations.

Why is it the case that our model, which is only trained on 2\% of the amount of training data of the original paper, gets very close with respect to nouns, but not with respect other part of speech tags, such as verbs, adjectives, and adverbs? It could be that the LSTM benefits from the additional 99 billion tokens to capture syntactic properties needed to disambiguate verbs, whereas this is less relevant for the disambiguation of nouns.

\subsection{Effect of the Amount of Training Data}
\label{sec:effect-of-data}

\begin{table}[t]
\small
\centering
\begin{tabular}{lc}
\toprule
Data size (\%) & Perplexity \\
\midrule
1              & 286.3           \\
10             & 50.4            \\
25             & 31.8            \\ 
\midrule
\newcite{kuchaiev.ginsburg2017} & 28.1 \\
\bottomrule
\end{tabular}
\caption{Effect of training set size on language modeling perplexity}
\label{tab:data-size-perplexity}
\end{table}

Since unannotated text is abundant, it is very tempting to use more and more data to train language models, hoping better embeddings translate into improved WSD performance.
The fact that \citeauthor{Yuan2016} used a 100-billion-word corpus only reinforces this intuition.
In this section, we empirically evaluate the effectiveness of data by varying the size of data used to train LSTM models and measure corresponding WSD performance.

We use a separate development set of 20K sentences and select random sentences from the rest of GigaWord to create a training set.
The size of the training set was set at 1\%, 10\%, 25\% and 100\% of the corpus.
For language modeling task, we use our second-largest model ($h=512, p=128$) and for WSD, we evaluate on SemEval13 with MFS fallback.

Table~\ref{tab:data-size-perplexity} reports the change of perplexity as more data is added.
To put the results in perspective, we compare to the reported
result of \newcite{kuchaiev.ginsburg2017} which is among the state-of-the-art in
language modeling.
The results verify that adding training data improves the quality of the language model.

\begin{figure}
\centering
\includegraphics[width=0.4\textwidth]{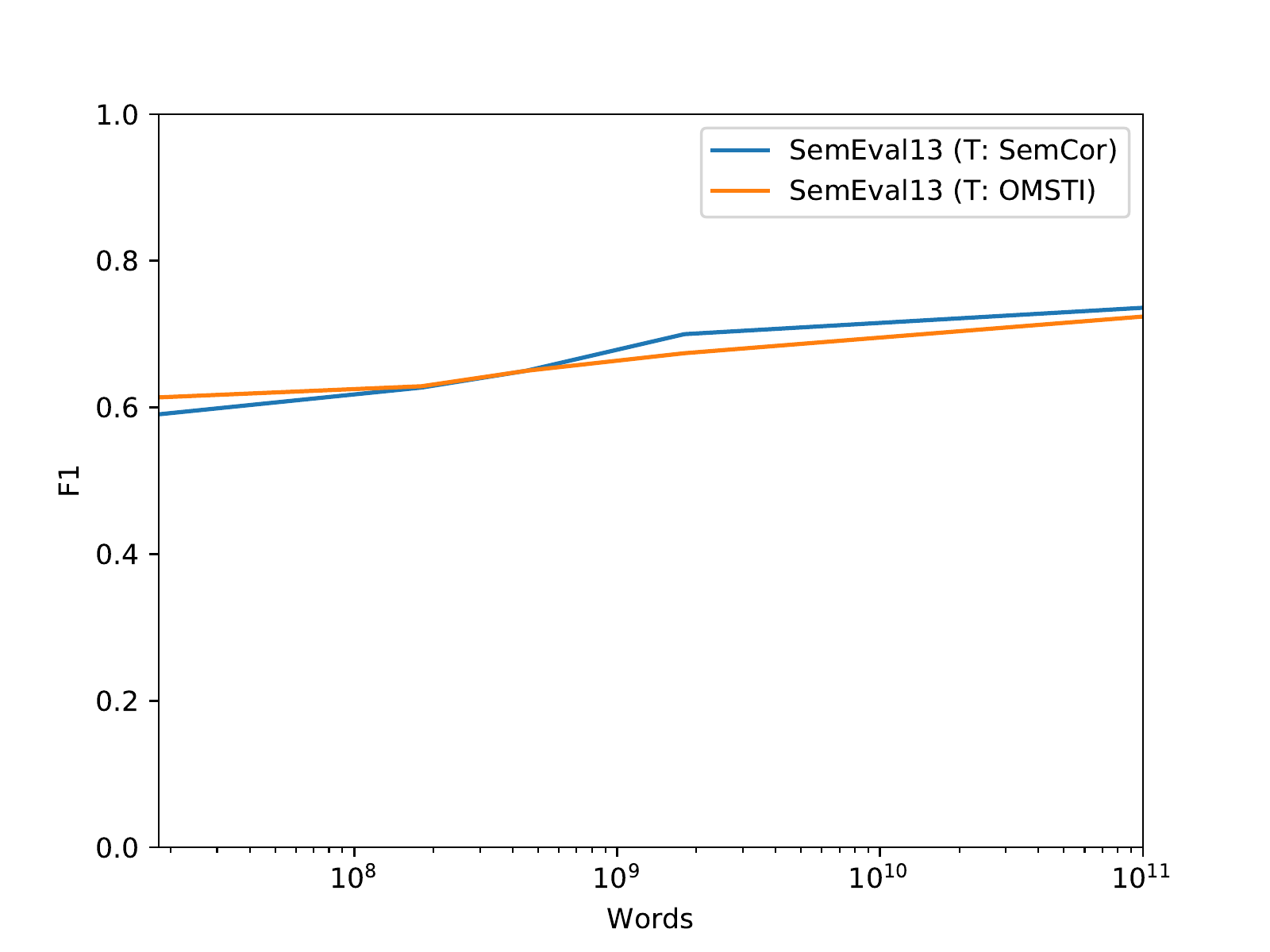}
\caption{The effect of the size of unannotated corpus on WSD performance. Notice that the horizontal axis is in log scale.}
\label{fig:data_size_vs_performance}
\end{figure}

Figure~\ref{fig:data_size_vs_performance} shows the effect of unannotated data
on WSD performance.
The data points at 100 billion words
correspond to \citeauthor{Yuan2016}'s reported results.
As can be expected, a bigger corpus leads to more meaningful context vectors and
therefore higher performance on WSD. However, the amount of added data for 1\%
of improvement in F1 grows exponentially fast (notice that the horizontal axis
is in log scale). Extrapolating from this graph, to get a performance of 0.8 F1
by adding more unannotated data, one would need a corpus of $10^{12}$ words.

\subsection{Effect of Language Model Capacity}
\label{sec:capacity}

One might suspect that the models in Section~\ref{sec:effect-of-data} are not big enough to absorb the added data. We argue that this is not the case.

\begin{figure}
\centering
\includegraphics[width=0.4\textwidth]{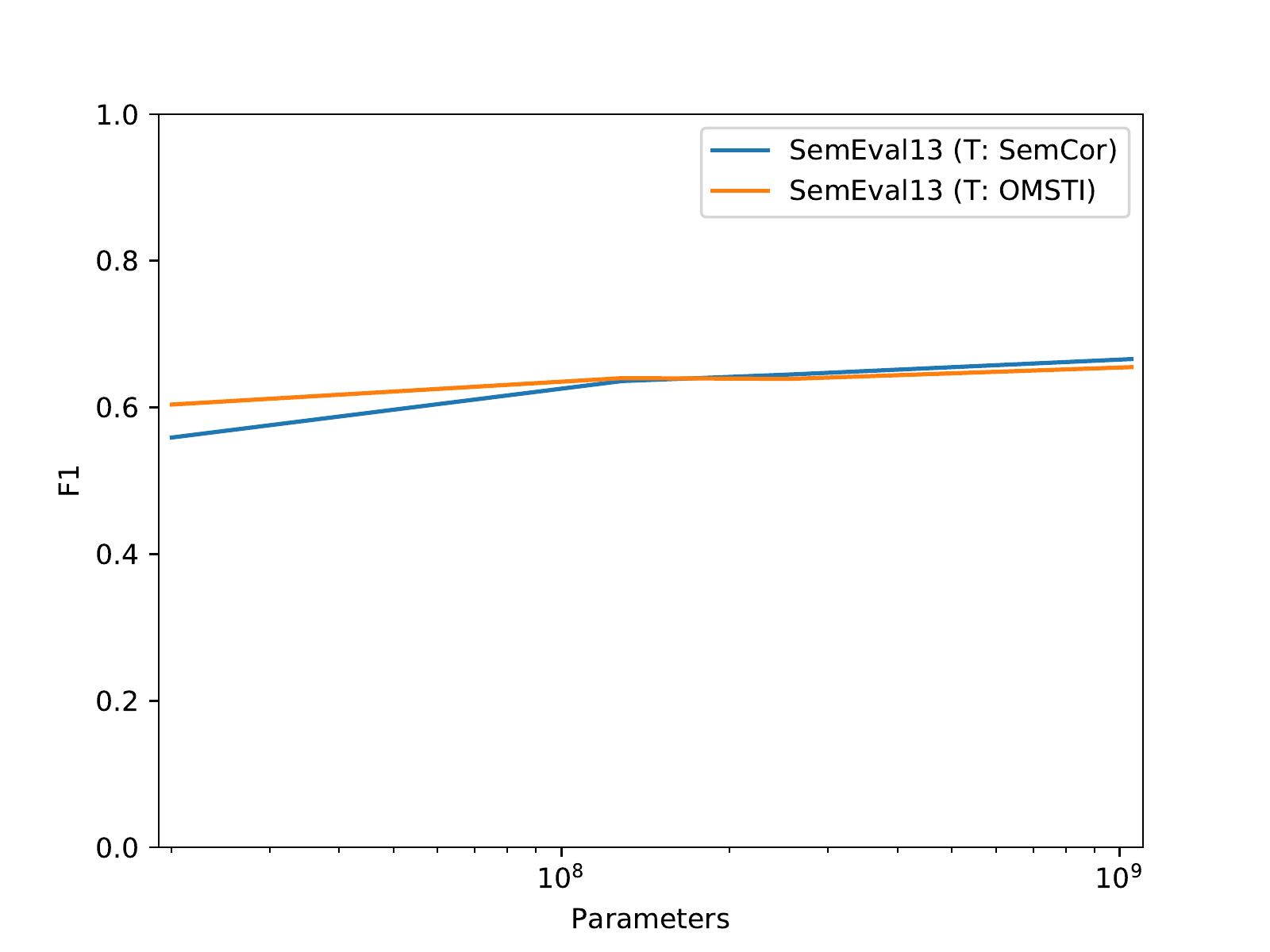}
\caption{The effect of LSTM model capacity on WSD performance. Notice that the horizontal axis is in log scale.}
\label{fig:capacity_vs_performance}
\end{figure}

Figure~\ref{fig:capacity_vs_performance} plots WSD performance against number of parameters in the LSTM model.
We train our LSTM models on 100\% of GigaWord corpus and evaluate it against SemEval13 with MFS fallback.
It is clear that increasing capacity brings only negligible improvement (notice that the number of parameters is plotted in log scale).
To improve WSD performance further, one should consider qualitatively different models instead of adding more data and computing power.

\subsection{Stability}
\label{sec:stability}

\newcite{reimers.gurevych2017} reported the performance of LSTM models on five NLP tasks.
Their results show that it is crucial to report the distribution of test scores instead of only one score as they might lead to opposite conclusions.
As pointed out in the beginning of Section~\ref{sec:results}, our biggest models take months to train, making training multiple versions of them impractical.
However, we report results of smaller models as a way to gauge the extent of random variation (Table~\ref{tab:stability}).
When trained with different random seeds, the models show slight difference in performance but it does not affect our conclusions.

\begin{table}[t]
\centering
\small
\begin{tabular}{lcc}
\toprule
 & Mean F1 & Std. \\
\midrule
$h=100,p=10$ (T: SemCor) & 0.559 & 0.008 \\
$h=100,p=10$ (T: OMSTI)  & 0.604 & 0.005 \\
$h=256,p=64$ (T: SemCor) & 0.636 & 0.005 \\
$h=256,p=64$ (T: OMSTI)  & 0.640 & 0.003 \\
\bottomrule
\end{tabular}
\caption{Performance of our models on SemEval13 when trained with different random seeds.}
\label{tab:stability}
\end{table}

\section{Conclusions}
\label{sec:conclusions}

In the current research, we do not aim at reproducing the same results reported in \newcite{Yuan2016} because we do not have access to the amount of data and computational power that Google researchers do. Instead, we hope to build a software package that can be deployed to the computing facilities of common research universities and publish a set of high-quality models to facilitate WSD research.

The analysis of our reimplemented model reveals the dependence of the model on a most-frequent-sense back-off strategy on one hand and relatively good performance in less-frequent-sense test instances on the other hand. We publish our models and source code so that researchers can perform further analysis.

We make a rather surprising observation that we do not need a huge unannotated corpus to achieve state-of-the-art WSD performance.
We train our LSTM language model on GigaWord which is two orders of magnitude smaller than \citeauthor{Yuan2016}'s corpus and got performance higher than many state-of-the-art systems on Senseval2 and SemEval13. Deeper analysis reveals that adding more words is subject to diminishing returns and increasing model capacity is not a viable solution.


We publish our code with detailed replication instructions at: \url{https://github.com/cltl/wsd-dynamic-sense-vector}
and our trained models at: \url{http://kyoto.let.vu.nl/~minh/wsd/}.\end{filecontents*}

\begin{filecontents*}{acl2017.sty}
\newif\ifacl@hyperref
\DeclareOption{hyperref}{\acl@hyperreftrue}
\DeclareOption{nohyperref}{\acl@hyperreffalse}
\ExecuteOptions{hyperref} 
\ProcessOptions\relax
\ifacl@hyperref
  \RequirePackage{hyperref}
  \usepackage{xcolor}		
  \definecolor{darkblue}{rgb}{0, 0, 0.5}
  \hypersetup{colorlinks=true,citecolor=darkblue, linkcolor=darkblue, urlcolor=darkblue}
\else
  \def\href#1#2{{#2}}
\fi

\typeout{Conference Style for ACL 2017}

%

\newcommand{\Thanks}[1]{\thanks{\ #1}}

\setlength{\paperwidth}{21cm}   
\setlength{\paperheight}{29.7cm}
\setlength\topmargin{-0.5cm}
\setlength\oddsidemargin{0cm}
\setlength\textheight{24.7cm}
\setlength\textwidth{16.0cm}
\setlength\columnsep{0.6cm}
\newlength\titlebox
\setlength\titlebox{5cm}
\setlength\headheight{5pt}
\setlength\headsep{0pt}
\thispagestyle{empty}
\pagestyle{empty}

\flushbottom \twocolumn \sloppy

\def\addcontentsline#1#2#3{}

\newif\ifaclfinal
\aclfinalfalse
\def\aclfinalcopy{\global\aclfinaltrue}

\newcommand{\@EveryShipoutACL@Hook}{}
\newcommand{\@EveryShipoutACL@AtNextHook}{}
\newcommand*{\EveryShipoutACL}[1]
   {\g@addto@macro\@EveryShipoutACL@Hook{#1}}
\newcommand*{\AtNextShipoutACL@}[1]
   {\g@addto@macro\@EveryShipoutACL@AtNextHook{#1}}
\newcommand{\@EveryShipoutACL@Shipout}{%
   \afterassignment\@EveryShipoutACL@Test
   \global\setbox\@cclv= %
   }
\newcommand{\@EveryShipoutACL@Test}{%
   \ifvoid\@cclv\relax
      \aftergroup\@EveryShipoutACL@Output
   \else
      \@EveryShipoutACL@Output
   \fi%
   }
\newcommand{\@EveryShipoutACL@Output}{%
   \@EveryShipoutACL@Hook%
   \@EveryShipoutACL@AtNextHook%
      \gdef\@EveryShipoutACL@AtNextHook{}%
   \@EveryShipoutACL@Org@Shipout\box\@cclv%
   }
\newcommand{\@EveryShipoutACL@Org@Shipout}{}
\newcommand*{\@EveryShipoutACL@Init}{%
   \message{ABD: EveryShipout initializing macros}%
   \let\@EveryShipoutACL@Org@Shipout\shipout
   \let\shipout\@EveryShipoutACL@Shipout
   }
\AtBeginDocument{\@EveryShipoutACL@Init}

\newcommand\LenToUnit[1]{#1\@gobble}
\newcommand\AtPageUpperLeft[1]{%
  \begingroup
    \@tempdima=0pt\relax\@tempdimb=\ESO@yoffsetI\relax
    \put(\LenToUnit{\@tempdima},\LenToUnit{\@tempdimb}){#1}%
  \endgroup
}
\newcommand\AtPageLowerLeft[1]{\AtPageUpperLeft{%
  \put(0,\LenToUnit{-\paperheight}){#1}}}
\newcommand\AtPageCenter[1]{\AtPageUpperLeft{%
  \put(\LenToUnit{.5\paperwidth},\LenToUnit{-.5\paperheight}){#1}}}
\newcommand\AtPageLowerCenter[1]{\AtPageUpperLeft{%
  \put(\LenToUnit{.5\paperwidth},\LenToUnit{-\paperheight}){#1}}}%
\newcommand\AtPageLowishCenter[1]{\AtPageUpperLeft{%
  \put(\LenToUnit{.5\paperwidth},\LenToUnit{-.96\paperheight}){#1}}}
\newcommand\AtTextUpperLeft[1]{%
  \begingroup
    \setlength\@tempdima{1in}%
    \advance\@tempdima\oddsidemargin%
    \@tempdimb=\ESO@yoffsetI\relax\advance\@tempdimb-1in\relax%
    \advance\@tempdimb-\topmargin%
    \advance\@tempdimb-\headheight\advance\@tempdimb-\headsep%
    \put(\LenToUnit{\@tempdima},\LenToUnit{\@tempdimb}){#1}%
  \endgroup
}
\newcommand\AtTextLowerLeft[1]{\AtTextUpperLeft{%
  \put(0,\LenToUnit{-\textheight}){#1}}}
\newcommand\AtTextCenter[1]{\AtTextUpperLeft{%
  \put(\LenToUnit{.5\textwidth},\LenToUnit{-.5\textheight}){#1}}}
\newcommand{\ESO@HookI}{} \newcommand{\ESO@HookII}{}
\newcommand{\ESO@HookIII}{}
\newcommand{\AddToShipoutPicture}{%
  \@ifstar{\g@addto@macro\ESO@HookII}{\g@addto@macro\ESO@HookI}}
\newcommand{\ClearShipoutPicture}{\global\let\ESO@HookI\@empty}
\newcommand{\@ShipoutPicture}{%
  \bgroup
    \@tempswafalse%
    \ifx\ESO@HookI\@empty\else\@tempswatrue\fi%
    \ifx\ESO@HookII\@empty\else\@tempswatrue\fi%
    \ifx\ESO@HookIII\@empty\else\@tempswatrue\fi%
    \if@tempswa%
      \@tempdima=1in\@tempdimb=-\@tempdima%
      \advance\@tempdimb\ESO@yoffsetI%
      \unitlength=1pt%
      \global\setbox\@cclv\vbox{%
        \vbox{\let\protect\relax
          \pictur@(0,0)(\strip@pt\@tempdima,\strip@pt\@tempdimb)%
            \ESO@HookIII\ESO@HookI\ESO@HookII%
            \global\let\ESO@HookII\@empty%
          \endpicture}%
          \nointerlineskip%
        \box\@cclv}%
    \fi
  \egroup
}
\EveryShipoutACL{\@ShipoutPicture}
\newif\ifESO@dvips\ESO@dvipsfalse
\newif\ifESO@grid\ESO@gridfalse
\newif\ifESO@texcoord\ESO@texcoordfalse
\newcommand*\ESO@griddelta{}\newcommand*\ESO@griddeltaY{}
\newcommand*\ESO@gridDelta{}\newcommand*\ESO@gridDeltaY{}
\newcommand*\ESO@yoffsetI{}\newcommand*\ESO@yoffsetII{}
\ifESO@texcoord
  \def\ESO@yoffsetI{0pt}\def\ESO@yoffsetII{-\paperheight}
  \edef\ESO@griddeltaY{-\ESO@griddelta}\edef\ESO@gridDeltaY{-\ESO@gridDelta}
\else
  \def\ESO@yoffsetI{\paperheight}\def\ESO@yoffsetII{0pt}
  \edef\ESO@griddeltaY{\ESO@griddelta}\edef\ESO@gridDeltaY{\ESO@gridDelta}
\fi


\font\naaclhv  = phvb at 8pt


\newbox\aclrulerbox
\newcount\aclrulercount
\newdimen\aclruleroffset
\newdimen\cv@lineheight
\newdimen\cv@boxheight
\newbox\cv@tmpbox
\newcount\cv@refno
\newcount\cv@tot
\newcount\cv@tmpc@ \newcount\cv@tmpc
\def\fillzeros[#1]#2{\cv@tmpc@=#2\relax\ifnum\cv@tmpc@<0\cv@tmpc@=-\cv@tmpc@\fi
\cv@tmpc=1 %
\loop\ifnum\cv@tmpc@<10 \else \divide\cv@tmpc@ by 10 \advance\cv@tmpc by 1 \fi
   \ifnum\cv@tmpc@=10\relax\cv@tmpc@=11\relax\fi \ifnum\cv@tmpc@>10 \repeat
\ifnum#2<0\advance\cv@tmpc1\relax-\fi
\loop\ifnum\cv@tmpc<#1\relax0\advance\cv@tmpc1\relax\fi \ifnum\cv@tmpc<#1 \repeat
\cv@tmpc@=#2\relax\ifnum\cv@tmpc@<0\cv@tmpc@=-\cv@tmpc@\fi \relax\the\cv@tmpc@}%
\def\makevruler[#1][#2][#3][#4][#5]{\begingroup\offinterlineskip
\textheight=#5\vbadness=10000\vfuzz=120ex\overfullrule=0pt%
\global\setbox\aclrulerbox=\vbox to \textheight{%
{\parskip=0pt\hfuzz=150em\cv@boxheight=\textheight
\cv@lineheight=#1\global\aclrulercount=#2%
\cv@tot\cv@boxheight\divide\cv@tot\cv@lineheight\advance\cv@tot2%
\cv@refno1\vskip-\cv@lineheight\vskip1ex%
\loop\setbox\cv@tmpbox=\hbox to0cm{{\naaclhv\hfil\fillzeros[#4]\aclrulercount}}%
\ht\cv@tmpbox\cv@lineheight\dp\cv@tmpbox0pt\box\cv@tmpbox\break
\advance\cv@refno1\global\advance\aclrulercount#3\relax
\ifnum\cv@refno<\cv@tot\repeat}}\endgroup}%

\def\aclpaperid{***}
\def\confidential{ACL 2017 Submission~\aclpaperid.  Confidential Review Copy.  DO NOT DISTRIBUTE.}

\def\aclruler#1{\makevruler[14.17pt][#1][1][3][\textheight]\usebox{\aclrulerbox}}
\def\leftoffset{-2.1cm} 
\def\rightoffset{17.5cm} 
\ifaclfinal\else\pagenumbering{arabic}
\AddToShipoutPicture{%
\ifaclfinal\else
\AtPageLowishCenter{\thepage}
\aclruleroffset=\textheight
\advance\aclruleroffset4pt
  \AtTextUpperLeft{%
    \put(\LenToUnit{\leftoffset},\LenToUnit{-\aclruleroffset}){
      \aclruler{\aclrulercount}}
    \put(\LenToUnit{\rightoffset},\LenToUnit{-\aclruleroffset}){
      \aclruler{\aclrulercount}}
  }
  \AtTextUpperLeft{
    \put(0,\LenToUnit{1cm}){\parbox{\textwidth}{\centering\naaclhv\confidential}}
  }
\fi
}




\newcommand\outauthor{
    \begin{tabular}[t]{c}
	\ifaclfinal
	     \bf\@author
	\else
     		\bf Anonymous ACL submission
	\fi
    \end{tabular}}

\ifaclfinal
\else
	\addtolength\titlebox{.25in}
\fi
\def\maketitle{\par
 \begingroup
   \def\thefootnote{\fnsymbol{footnote}}
   \def\@makefnmark{\hbox to 0pt{$^{\@thefnmark}$\hss}}
   \twocolumn[\@maketitle] \@thanks
 \endgroup
 \setcounter{footnote}{0}
 \let\maketitle\relax \let\@maketitle\relax
 \gdef\@thanks{}\gdef\@author{}\gdef\@title{}\let\thanks\relax}
\def\@maketitle{\vbox to \titlebox{\hsize\textwidth
 \linewidth\hsize \vskip 0.125in minus 0.125in \centering
 {\Large\bf \@title \par} \vskip 0.2in plus 1fil minus 0.1in
 {\def\and{\unskip\enspace{\rm and}\enspace}%
  \def\And{\end{tabular}\hss \egroup \hskip 1in plus 2fil
           \hbox to 0pt\bgroup\hss \begin{tabular}[t]{c}\bf}%
  \def\AND{\end{tabular}\hss\egroup \hfil\hfil\egroup
          \vskip 0.25in plus 1fil minus 0.125in
           \hbox to \linewidth\bgroup\large \hfil\hfil
             \hbox to 0pt\bgroup\hss \begin{tabular}[t]{c}\bf}
  \hbox to \linewidth\bgroup\large \hfil\hfil
    \hbox to 0pt\bgroup\hss
	\outauthor
   \hss\egroup
    \hfil\hfil\egroup}
  \vskip 0.3in plus 2fil minus 0.1in
}}

\renewenvironment{abstract}%
		 {\centerline{\large\bf Abstract}%
		  \begin{list}{}%
		     {\setlength{\rightmargin}{0.6cm}%
		      \setlength{\leftmargin}{0.6cm}}%
		   \item[]\ignorespaces}%
		 {\unskip\end{list}}

\RequirePackage{natbib}
\renewcommand\cite{\citep}	
\newcommand\shortcite{\citeyearpar}
\newcommand\newcite{\citet}	


\def\@up#1{\raise.2ex\hbox{#1}}

\def\thebibliography#1{\vskip\parskip%
\vskip\baselineskip%
\def\baselinestretch{1}%
\ifx\@currsize\normalsize\@normalsize\else\@currsize\fi%
\vskip-\parskip%
\vskip-\baselineskip%
\section*{References\@mkboth
 {References}{References}}\list
 {}{\setlength{\labelwidth}{0pt}\setlength{\leftmargin}{\parindent}
 \setlength{\itemindent}{-\parindent}}
 \def\newblock{\hskip .11em plus .33em minus -.07em}
 \sloppy\clubpenalty4000\widowpenalty4000
 \sfcode`\.=1000\relax}
\let\endthebibliography=\endlist

\def\thesourcebibliography#1{\vskip\parskip%
\vskip\baselineskip%
\def\baselinestretch{1}%
\ifx\@currsize\normalsize\@normalsize\else\@currsize\fi%
\vskip-\parskip%
\vskip-\baselineskip%
\section*{Sources of Attested Examples\@mkboth
 {Sources of Attested Examples}{Sources of Attested Examples}}\list
 {}{\setlength{\labelwidth}{0pt}\setlength{\leftmargin}{\parindent}
 \setlength{\itemindent}{-\parindent}}
 \def\newblock{\hskip .11em plus .33em minus -.07em}
 \sloppy\clubpenalty4000\widowpenalty4000
 \sfcode`\.=1000\relax}
\let\endthesourcebibliography=\endlist

\def\section{\@startsection {section}{1}{\z@}{-2.0ex plus
    -0.5ex minus -.2ex}{1.5ex plus 0.3ex minus .2ex}{\large\bf\raggedright}}
\def\subsection{\@startsection{subsection}{2}{\z@}{-1.8ex plus
    -0.5ex minus -.2ex}{0.8ex plus .2ex}{\normalsize\bf\raggedright}}
\def\subsubsection{\@startsection{subsubsection}{3}{\z@}{-1.5ex plus
   -0.5ex minus -.2ex}{0.5ex plus .2ex}{\normalsize\bf\raggedright}}
\def\paragraph{\@startsection{paragraph}{4}{\z@}{1.5ex plus
   0.5ex minus .2ex}{-1em}{\normalsize\bf}}
\def\subparagraph{\@startsection{subparagraph}{5}{\parindent}{1.5ex plus
   0.5ex minus .2ex}{-1em}{\normalsize\bf}}

\footnotesep 6.65pt %
\skip\footins 9pt plus 4pt minus 2pt
\def\footnoterule{\kern-3pt \hrule width 5pc \kern 2.6pt }
\setcounter{footnote}{0}

\parindent 1em
\topsep 4pt plus 1pt minus 2pt
\partopsep 1pt plus 0.5pt minus 0.5pt
\itemsep 2pt plus 1pt minus 0.5pt
\parsep 2pt plus 1pt minus 0.5pt

\leftmargin 2em \leftmargini\leftmargin \leftmarginii 2em
\leftmarginiii 1.5em \leftmarginiv 1.0em \leftmarginv .5em \leftmarginvi .5em
\labelwidth\leftmargini\advance\labelwidth-\labelsep \labelsep 5pt

\def\@listi{\leftmargin\leftmargini}
\def\@listii{\leftmargin\leftmarginii
   \labelwidth\leftmarginii\advance\labelwidth-\labelsep
   \topsep 2pt plus 1pt minus 0.5pt
   \parsep 1pt plus 0.5pt minus 0.5pt
   \itemsep \parsep}
\def\@listiii{\leftmargin\leftmarginiii
    \labelwidth\leftmarginiii\advance\labelwidth-\labelsep
    \topsep 1pt plus 0.5pt minus 0.5pt
    \parsep \z@ \partopsep 0.5pt plus 0pt minus 0.5pt
    \itemsep \topsep}
\def\@listiv{\leftmargin\leftmarginiv
     \labelwidth\leftmarginiv\advance\labelwidth-\labelsep}
\def\@listv{\leftmargin\leftmarginv
     \labelwidth\leftmarginv\advance\labelwidth-\labelsep}
\def\@listvi{\leftmargin\leftmarginvi
     \labelwidth\leftmarginvi\advance\labelwidth-\labelsep}

\abovedisplayskip 7pt plus2pt minus5pt%
\belowdisplayskip \abovedisplayskip
\abovedisplayshortskip  0pt plus3pt%
\belowdisplayshortskip  4pt plus3pt minus3pt%

\def\@normalsize{\@setsize\normalsize{11pt}\xpt\@xpt}
\def\small{\@setsize\small{10pt}\ixpt\@ixpt}
\def\footnotesize{\@setsize\footnotesize{10pt}\ixpt\@ixpt}
\def\scriptsize{\@setsize\scriptsize{8pt}\viipt\@viipt}
\def\tiny{\@setsize\tiny{7pt}\vipt\@vipt}
\def\large{\@setsize\large{14pt}\xiipt\@xiipt}
\def\Large{\@setsize\Large{16pt}\xivpt\@xivpt}
\def\LARGE{\@setsize\LARGE{20pt}\xviipt\@xviipt}
\def\huge{\@setsize\huge{23pt}\xxpt\@xxpt}
\def\Huge{\@setsize\Huge{28pt}\xxvpt\@xxvpt}

\section*{Acknowledgements}

The research for this paper was supported by the Netherlands Organisation for Scientific Research (NWO) via the Spinoza-prize Vossen projects (SPI 30-673, 2014-2019).
We thank the support of the NWO project scilens (\url{https://projects.cwi.nl/scilens}) for providing the hardware to run some of the experiments.
Experiments were also carried out on the Dutch national e-infrastructure with the support of SURF Cooperative.
We would like to thank our friends and colleagues Piek Vossen and Antske Fokkens 
for many useful comments and discussions.

\bibliographystyle{acl_natbib}

\begin{thebibliography}{}
\expandafter\ifx\csname natexlab\endcsname\relax\def\natexlab#1{#1}\fi

\bibitem[{Abadi et~al.(2015)Abadi, Agarwal, Barham, Brevdo, Chen, Citro,
  Corrado, Davis, Dean, Devin, Ghemawat, Goodfellow, Harp, Irving, Isard, Jia,
  Kaiser, Kudlur, Levenberg, Man, Monga, Moore, Murray, Shlens, Steiner,
  Sutskever, Tucker, Vanhoucke, Vasudevan, Vinyals, Warden, Wicke, Yu, and
  Zheng}]{Abadi2015}
Martin Abadi, Ashish Agarwal, Paul Barham, Eugene Brevdo, Zhifeng Chen, Craig
  Citro, Greg Corrado, Andy Davis, Jeffrey Dean, Matthieu Devin, Sanjay
  Ghemawat, Ian Goodfellow, Andrew Harp, Geoffrey Irving, Michael Isard,
  Yangqing Jia, Lukasz Kaiser, Manjunath Kudlur, Josh Levenberg, Dan Man, Rajat
  Monga, Sherry Moore, Derek Murray, Jon Shlens, Benoit Steiner, Ilya
  Sutskever, Paul Tucker, Vincent Vanhoucke, Vijay Vasudevan, Oriol Vinyals,
  Pete Warden, Martin Wicke, Yuan Yu, and Xiaoqiang Zheng. 2015.
\newblock {TensorFlow: Large-Scale Machine Learning on Heterogeneous
  Distributed Systems}.
\newblock {\em arXiv:1603.04467v2\/} page~19.

\bibitem[{Agirre et~al.(2014)Agirre, de~Lacalle, and Soroa}]{agirre2014random}
Eneko Agirre, Oier~Lopez de~Lacalle, and Aitor Soroa. 2014.
\newblock {R}andom {W}alks for {K}nowledge-{B}ased {W}ord {S}ense
  {D}isambiguation.
\newblock {\em Computational Linguistics\/} 40(1):57--84.

\bibitem[{Dyer et~al.(2015)Dyer, Ballesteros, Ling, Matthews, and
  Smith}]{Dyer2015}
Chris Dyer, Miguel Ballesteros, Wang Ling, Austin Matthews, and Noah~A Smith.
  2015.
\newblock {Transition-Based Dependency Parsing with Stack Long Short-Term
  Memory}.
\newblock In {\em ACL 2015\/}. pages 334--343.

\bibitem[{Fellbaum(1998)}]{Fellbaum-WordNet-1998}
Christiane Fellbaum, editor. 1998.
\newblock {\em WordNet An Electronic Lexical Database\/}.
\newblock The MIT Press, Cambridge, MA ; London.

\bibitem[{Francis and Kucera(1979)}]{francis1979brown}
Winthrop~Nelson Francis and Henry Kucera. 1979.
\newblock Brown corpus manual.
\newblock {\em Brown University\/} .

\bibitem[{Goodfellow et~al.(2016)Goodfellow, Bengio, and
  Courville}]{Goodfellow-et-al-2016}
Ian Goodfellow, Yoshua Bengio, and Aaron Courville. 2016.
\newblock {\em Deep Learning\/}.
\newblock MIT Press.

\bibitem[{He et~al.(2017)He, Lee, Lewis, and Zettlemoyer}]{He2017}
Luheng He, Kenton Lee, Mike Lewis, and Luke Zettlemoyer. 2017.
\newblock {Deep Semantic Role Labeling : What Works and What’s Next}.
\newblock {\em Acl2017\/} .

\bibitem[{Hochreiter and Schmidhuber(1997)}]{hochreiter.schmidhuber97}
Sepp Hochreiter and Jürgen Schmidhuber. 1997.
\newblock {Long short-term memory}.
\newblock {\em Neural computation\/} 9(8):1735--1780.

\bibitem[{Iacobacci et~al.(2016)Iacobacci, Pilehvar, and Navigli}]{P16-1085}
Ignacio Iacobacci, Mohammad~Taher Pilehvar, and Roberto Navigli. 2016.
\newblock \href{https://doi.org/10.18653/v1/P16-1085}{Embeddings for word sense
  disambiguation: An evaluation study}.
\newblock In {\em Proceedings of the 54th Annual Meeting of the Association for
  Computational Linguistics (Volume 1: Long Papers)\/}. Association for
  Computational Linguistics, pages 897--907.
\newblock
  \href{https://doi.org/10.18653/v1/P16-1085}{https://doi.org/10.18653/v1/P16-1085}.

\bibitem[{Jean et~al.(2014)Jean, Cho, Memisevic, and Bengio}]{Jean2014}
S{\'{e}}bastien Jean, Kyunghyun Cho, Roland Memisevic, and Yoshua Bengio. 2014.
\newblock \href{https://doi.org/10.3115/v1/P15-1001}{{On Using Very Large
  Target Vocabulary for Neural Machine Translation}} pages 1--10.
\newblock
  \href{https://doi.org/10.3115/v1/P15-1001}{https://doi.org/10.3115/v1/P15-1001}.

\bibitem[{Kuchaiev and Ginsburg(2017)}]{kuchaiev.ginsburg2017}
Oleksii Kuchaiev and Boris Ginsburg. 2017.
\newblock {Factorization Tricks for LSMT Networks}.
\newblock In {\em ICLR 2017\/}. pages 1--5.

\bibitem[{Miller et~al.(1993)Miller, Leacock, Tengi, and
  T.~Bunker}]{miller1993semantic}
George Miller, Claudia Leacock, Randee Tengi, and Ross T.~Bunker. 1993.
\newblock A semantic concordance.
\newblock In {\em HUMAN LANGUAGE TECHNOLOGY: Proceedings of a Workshop Held at
  Plainsboro, New Jersey, March 21-24, 1993\/}.

\bibitem[{Navigli(2009)}]{navigli2009word}
Roberto Navigli. 2009.
\newblock Word sense disambiguation: A survey.
\newblock {\em ACM Computing Surveys (CSUR)\/} 41(2):10.

\bibitem[{Navigli et~al.(2013)Navigli, Jurgens, and
  Vannella}]{navigli-jurgens-vannella:2013:SemEval-2013}
Roberto Navigli, David Jurgens, and Daniele Vannella. 2013.
\newblock Sem{E}val-2013 task 12: {M}ultilingual {W}ord {S}ense
  {D}isambiguation.
\newblock In {\em Second Joint Conference on Lexical and Computational
  Semantics (*SEM), Volume 2: Proceedings of the Seventh International Workshop
  on Semantic Evaluation (SemEval 2013)\/}. Association for Computational
  Linguistics, Atlanta, Georgia, USA, pages 222--231.

\bibitem[{Palmer et~al.(2001)Palmer, Fellbaum, Cotton, Delfs, and
  Dang}]{palmer-EtAl:2001:SENSEVAL}
Martha Palmer, Christiane Fellbaum, Scott Cotton, Lauren Delfs, and Hoa~Trang
  Dang. 2001.
\newblock English {T}asks: {A}ll-{W}ords and {V}erb {L}exical {S}ample.
\newblock In {\em Proceedings of the Second International Workshop on
  Evaluating Word Sense Disambiguation Systems (SENSEVAL-2)\/}. Association for
  Computational Linguistics, Toulouse, France, pages 21--24.

\bibitem[{Postma et~al.(2016{\natexlab{a}})Postma, Izquierdo, Agirre, Rigau,
  and Vossen}]{POSTMA16.437}
Marten Postma, Ruben Izquierdo, Eneko Agirre, German Rigau, and Piek Vossen.
  2016{\natexlab{a}}.
\newblock Addressing the {MFS} {B}ias in {WSD} systems.
\newblock In Nicoletta Calzolari~(Conference Chair), Khalid Choukri, Thierry
  Declerck, Sara Goggi, Marko Grobelnik, Bente Maegaard, Joseph Mariani, Helene
  Mazo, Asuncion Moreno, Jan Odijk, and Stelios Piperidis, editors, {\em
  Proceedings of the Tenth International Conference on Language Resources and
  Evaluation (LREC 2016)\/}. European Language Resources Association (ELRA),
  Paris, France.

\bibitem[{Postma et~al.(2016{\natexlab{b}})Postma, Izquierdo~Bevia, and
  Vossen}]{C16-1330}
Marten Postma, Ruben Izquierdo~Bevia, and Piek Vossen. 2016{\natexlab{b}}.
\newblock More is not always better: balancing sense distributions for
  all-words word sense disambiguation.
\newblock In {\em Proceedings of COLING 2016, the 26th International Conference
  on Computational Linguistics: Technical Papers\/}. The COLING 2016 Organizing
  Committee, pages 3496--3506.

\bibitem[{Raganato et~al.(2017)Raganato, Delli~Bovi, and Navigli}]{raganato}
Alessandro Raganato, Claudio Delli~Bovi, and Roberto Navigli. 2017.
\newblock Neural sequence learning models for word sense disambiguation.
\newblock In {\em Proceedings of the 2017 Conference on Empirical Methods in
  Natural Language Processing\/}. Association for Computational Linguistics,
  pages 1167--1178.

\bibitem[{Reimers and Gurevych(2017)}]{reimers.gurevych2017}
Nils Reimers and Iryna Gurevych. 2017.
\newblock {Reporting Score Distributions Makes a Difference: Performance Study
  of LSTM-networks for Sequence Tagging}.
\newblock In {\em EMNLP\/}. pages 338--348.

\bibitem[{Rothe and Sch{\"u}tze(2015)}]{P15-1173}
Sascha Rothe and Hinrich Sch{\"u}tze. 2015.
\newblock \href{https://doi.org/10.3115/v1/P15-1173}{Autoextend: Extending word
  embeddings to embeddings for synsets and lexemes}.
\newblock In {\em Proceedings of the 53rd Annual Meeting of the Association for
  Computational Linguistics and the 7th International Joint Conference on
  Natural Language Processing (Volume 1: Long Papers)\/}. Association for
  Computational Linguistics, pages 1793--1803.
\newblock
  \href{https://doi.org/10.3115/v1/P15-1173}{https://doi.org/10.3115/v1/P15-1173}.

\bibitem[{Sutskever et~al.(2014)Sutskever, Vinyals, and
  Le}]{sutskever2014sequence}
Ilya Sutskever, Oriol Vinyals, and Quoc~V Le. 2014.
\newblock {Sequence to sequence learning with neural networks}.
\newblock In {\em Advances in neural information processing systems\/}. pages
  3104--3112.

\bibitem[{Taghipour and Ng(2015)}]{K15-1037}
Kaveh Taghipour and Tou~Hwee Ng. 2015.
\newblock \href{https://doi.org/10.18653/v1/K15-1037}{One million sense-tagged
  instances for word sense disambiguation and induction}.
\newblock In {\em Proceedings of the Nineteenth Conference on Computational
  Natural Language Learning\/}. Association for Computational Linguistics,
  pages 338--344.
\newblock
  \href{https://doi.org/10.18653/v1/K15-1037}{https://doi.org/10.18653/v1/K15-1037}.

\bibitem[{Tripodi and Pelillo(2016)}]{tripodi2016game}
Rocco Tripodi and Marcello Pelillo. 2016.
\newblock A game-theoretic approach to word sense disambiguation.
\newblock {\em arXiv preprint arXiv:1606.07711\/} .

\bibitem[{Weissenborn et~al.(2015)Weissenborn, Hennig, Xu, and
  Uszkoreit}]{P15-1058}
Dirk Weissenborn, Leonhard Hennig, Feiyu Xu, and Hans Uszkoreit. 2015.
\newblock \href{https://doi.org/10.3115/v1/P15-1058}{Multi-objective
  optimization for the joint disambiguation of nouns and named entities}.
\newblock In {\em Proceedings of the 53rd Annual Meeting of the Association for
  Computational Linguistics and the 7th International Joint Conference on
  Natural Language Processing (Volume 1: Long Papers)\/}. Association for
  Computational Linguistics, pages 596--605.
\newblock
  \href{https://doi.org/10.3115/v1/P15-1058}{https://doi.org/10.3115/v1/P15-1058}.

\bibitem[{Yuan et~al.(2016)Yuan, Doherty, Richardson, Evans, and
  Altendorf}]{Yuan2016}
Dayu Yuan, Ryan Doherty, Julian Richardson, Colin Evans, and Eric Altendorf.
  2016.
\newblock {Semi-supervised Word Sense Disambiguation with Neural Language
  Models}.
\newblock In {\em Proceedings of COLING 2016, the 26th International Conference
  on Computational Linguistics: Technical Papers\/}. The COLING 2016 Organizing
  Committee, Osaka, Japan, pages 1374--1385.

\bibitem[{Zhong and Ng(2010)}]{Zhong:2010:MSW:1858933.1858947}
Zhi Zhong and Tou~Hwee Ng. 2010.
\newblock It makes sense: A wide-coverage word sense disambiguation system for
  free text.
\newblock In {\em Proceedings of the ACL 2010 System Demonstrations\/}.
  Association for Computational Linguistics, pages 78--83.

\end{thebibliography}

\end{document}